\documentclass[conference]{IEEEtran}
\IEEEoverridecommandlockouts
\usepackage{cite}
\usepackage{amsmath,amssymb,amsfonts}
\usepackage{algorithmic}
\usepackage{graphicx}
\usepackage{textcomp}
\usepackage{subfig}

\usepackage{todonotes}

\usepackage{acronym}

\usepackage{xcolor}
\usepackage{ifthen}
\usepackage{caption}
\usepackage{xspace}
\def\BibTeX{{\rm B\kern-.05em{\sc i\kern-.025em b}\kern-.08em
    T\kern-.1667em\lower.7ex\hbox{E}\kern-.125emX}}

\makeatletter 
\newcommand{\linebreakand}{%
  \end{@IEEEauthorhalign}
  \hfill\mbox{}\par
  \mbox{}\hfill\begin{@IEEEauthorhalign}
}
\makeatother 

\setuptodonotes{inline}

\newtheorem{thm:def}{Definition}
\newtheorem{thm:task}{Task}
\newtheorem{thm:action}{Action}

\newcommand\trained{\mathrel{\reflectbox{$\leadsto$}}}
\newcommand\trainedww{\trained_{\textit{ww}}}

\newcommand\applied{\lhd}
\newcommand\appliedww{\applied_{\textit{ww}}}
\newcommand\appliedrep{\applied_{\textit{rep}}}
\newcommand\appliedaux{\bullet_{\textit{aux}}}

\newcommand{\wrt}{w.r.t.\xspace}
\newcommand{\cf}{cf.\xspace}

\newcommand{\pnsr}{$PN_{SR}$\xspace}
\newcommand{\pnf}{$PN_{F}$\xspace}

\newcommand{\qquote}[1]{`#1'}

\let\acdef\acrodef
\renewcommand{\acrodef}[3]{\acdef{#1}[#2]{\emph{#3}}}

\acrodef{ml}{ML}{machine learning}
\acrodef{rl}{RL}{reinforcement learning}
\acrodef{sp}{SP}{state space}
\acrodef{bp}{BP}{back-propagation}

\acrodef{nn}{NN}{neural network} 
\acrodef{srnn}{SRNN}{self-replicating neural network}
\acrodef{sr}{SR}{self-replication}
\acrodef{pn}{PN}{particle network}
\acrodef{on}{ON}{organism network}
\acrodef{at}{AT}{auxilary task}
\acrodef{sa}{SA}{self-application}
\acrodef{dan}{DAN}{deep artificial neuron}
\acrodef{mnist}{MNIST}{MNIST-dataset}
\acrodef{c}{C}{cell}
\acrodef{gelu}{GELU}{Gaussian Error Linear Unit}

\title{Constructing Organism Networks from Collaborative Self-Replicators}

\begin{document}
\everypar{\looseness=-1}
\author{
    \centering
    \IEEEauthorblockN{
    \begin{minipage}{1\textwidth}
    \centering
    Steffen Illium, Maximilian Zorn, Cristian Lenta,\\
    Michael Kölle, Claudia Linnhoff-Popien, Thomas Gabor
    \end{minipage}
    }
    \IEEEauthorblockA{
    \footnote{test}\textit{Institute of Informatics}, \textit{LMU Munich}\\
    steffen.illium@ifi.lmu.de}
}
\maketitle

\begin{abstract}
    We introduce \aclp{on}, which function like a single \acl{nn} but are composed of several neural \aclp{pn}; while each \acl{pn} fulfils the role of a single weight application within the \acl{on}, it is also trained to self-replicate its own weights.
    As \aclp{on} feature vastly more parameters than simpler architectures, we perform our initial experiments on an arithmetic task as well as on simplified \acl{mnist} classification as a collective.
    We observe that individual \aclp{pn} tend to specialise in either of the tasks and that the ones fully specialised in the secondary task may be dropped from the network without hindering the computational accuracy of the primary task.
    This leads to the discovery of a novel pruning-strategy for sparse \aclp{nn}.
\end{abstract}

\section{Introduction}
\emph{Self-replication} (SR) has always been regarded one of the central aspects of life \cite{kauffman1993origins}. 
Recent literature has shown spiking interest in enabling \acfp{nn} to self-replicate (cf. \cite{chang2018neural,gabor2019self,randazzo2021fertile_self_replication}), mostly for the pure reason of creating more lifelike entities from \acp{nn}, which are currently the most prominent tools to encode complex behaviour in a way that is easily accessible to machine learning. 
Eventually, one may hope to lift some limitations imposed by current neural network architectures via means of \ac{sr} to decrease volatility in the training process or to create new means of information interchange between such networks.

Instances of \acp{srnn} have already been given additional tasks that are more directly useful to the system's designer than pure \ac{sr} --- a kind of ultimate purpose (cf. \cite{chang2018neural,gabor2021goals,randazzo2021fertile_self_replication}).
However, in all of these instances, an \ac{srnn} is expected to perform said task on its own or only with indirect support from other networks during training \cite{gabor2021goals}.
Furthermore, the tasks visited by Gabor et al.~\cite{gabor2021goals} can be described as very basic arithmetic operations.

Therefore, in this study, we built upon the work by Gabor et al.~\cite{gabor2021goals} in establishing singular \aclp{srnn} as members of a group of many networks that collaborate on learning to solve a joint \ac{at}.
Effectively, we define a structure we call an \acf{on}, which looks like any other neural network from the outside but on the inside is composed of various other \acp{nn} we call \acfp{pn}.
While each individual of these \acp{pn} is trained to self-replicate, the whole \ac{on} is trained to perform a \qquote{meaningful} (\emph{auxiliary}) task (like the simple addition of numbers or the classification of images).
Thus, we push the metaphor of self-replicating \acp{nn} one step further by having them behave in an (at least more) similar way to cells in multicellular biological specimen.

As the parameter space of the complete \acl{on} with all its \acp{pn} grows to large dimensions, we consider showing the feasibility of training such a structure to meaningful and useful behaviour as one main contribution of this paper, even though organism networks cannot quite compete with standard approaches for training efficiency on their collaborative \acl{at} yet. 
Most interestingly, we observe that, with a simpler problem like float addition, all \acp{pn} can be trained to be self-replicators while performing the \ac{at}; a more complicated task like classification of images in the \acf{mnist}, however, gives rise to fully autonomous specialisation within the \acp{pn}. 
This means that certain \acp{pn} tend to focus on \ac{sr} while others focus on the \acl{at}. 
Our experiments suggest that, when pruning these \acp{pn} from a comparable sized \ac{nn}, we can still retain almost all the \acl{on}'s external function (\ac{at}).
As a novel finding, we also observe better performance in comparison to a comparable network pruned by a global unstructured `\emph{l1-pruning}' method. 

The structure is as follows: We present a brief recap of relevant background in the field of \acl{srnn} in Section~\ref{sec:background}, before proposing the concept and architecture of \aclp{on} in Section~\ref{sec:Organism_Network_Intro}. 
Experiments with two different \aclp{at} of increasing difficulty and the investigation of resulting performances can be found in Sections~\ref{sec:exp_1_float_addition} \&~\ref{sec:exp_2_mnist_classification}. 
Finally, we discuss related work in Section~\ref{sec:related_work} and conclude our findings in Section~\ref{sec:conclusion_and_future_work}.

\section{Self-Replicating Networks}
\label{sec:background}

We follow Gabor et al.~\cite{gabor2019self} in the definition of self-replicating neural networks, of which we sum up shortly all relevant and adopted material. We will also briefly define our approach and notation for the underlying neural network basics.

\begin{thm:def}[\acl{nn}]\label{def:neural_network}
	A \emph{neural network} is a function $\mathcal{N}: \mathbb{R}^p \to \mathbb{R}^q$ with $p$ inputs and $q$ outputs. This function is defined via a graph made up of $r$ layers $L_1, ..., L_r$ where each layer $L_l$ consists of $|L_l|$ cells $C_1, ..., C_{|L_l|}$, which make up the graph's vertices, and each cell $C_{l,c}$ of the layer $L_l$ is connected to all cells of the previous layer, i.e., $C_{l-1,d}$ for $d = 1, ..., |L_{l-1}|$, via the graph's edges. Each edge of a cell $C_{l,c}$ is assigned an edge weight $E_{l,c,e} \in \mathbb{R}, e = 1, ..., |L_{l-1}|$. Given a fixed graph structure, the vector of all edge weights $\mathbf{w} = \overline{\mathcal{N}} = \langle E_{l,c,e} \rangle_{l=1,...,r, c = 1, ..., |L_l|, e = 1, ..., |L_{l-1}|}$ defines the network's functionality.
	
    A network's output given an input $\mathbf{x} \in \mathbb{R}^p$ is given via $\mathbf{y} = \mathcal{N}(\mathbf{x}) = \langle O(r,c) \rangle_{c=1,...,|L_r|} \in \mathbb{R}^q$ where $$O(l,c) = \begin{cases} x_c & \textit{ if } l = 0, \\ \sum_{i=1}^{|L_{l-1}|} E_{l,c,i} \cdot O(l-1,c) & \textit{ otherwise.}\end{cases}$$
\end{thm:def}

Note that Definition~\ref{def:neural_network} is not the most general way to define \aclp{nn} and especially our definition of $O$ already imposes linear activation as we use it for the basic \acp{srnn} in this paper. 
Further note that during all experiments, the network architecture remains fixed so that we can deduce a network's function using only its weight vector.
To better understand the internal structure of a neural network and how the function is derived from architecture and weights, please refer to Lecun et al.~\cite{lecun2015deep_learning} or the definitions by Gabor et al.~\cite{gabor2019self}, which match our formalism.

To allow for \ac{sr} on a formal level, we need to define the \ac{nn} in a way so that it can process all of its own weights. 
As the weight vector of an arbitrary network always contains at least as many weights as its input and output dimensions combined, it is rather difficult to implement a neural network that can process all its own parameters in a single activation step.
As discussed by Gabor et al.~\cite{gabor2019self}, it is possible to pass every weight parameter to the network one by one alongside with some meta-data regarding the weight parameter's position within the network (\wrt to layer, neuron, weight), which enables the \ac{nn} to process a vector as large as its own weight vector via several iterations.
We adopt this so called~\emph{weightwise reduction}~\cite{gabor2019self} for our research, which also allows for interactions between particles in future experiments.

An application of one network reduced in this way onto another network (or itself) is defined as follows:
\begin{thm:def}[weightwise application]\label{def:appliedww}
	Let $\mathcal{M},\mathcal{N},\mathcal{O}: \mathbb{R}^4 \to \mathbb{R}$ be neural networks. Let the weights of $\mathcal{M}$ be $\overline{\mathcal{M}} = \langle v_i \rangle_{0 \leq i < |\overline{\mathcal{M}}|}$. 
    A neural network $\mathcal{O}$ is the result of weightwise application of $\mathcal{N}$ onto $\mathcal{M}$, written $\mathcal{O} = \mathcal{N} \appliedww \mathcal{M}$, iff $$\overline{\mathcal{O}} = \langle \mathcal{N}(v_i, L(i), C(i), E(i))\rangle_{0 \leq i < |\overline{\mathcal{M}}|}$$ where $L(i)$ is the layer of the weight $i$, $C(i)$ is the cell the weight $i$ leads into and $E(i)$ is the positional number of edge weight $i$ among the weights of its cell.
\end{thm:def}



Gabor et al.~\cite{gabor2019self} define \ac{sr} as a network's ability to reproduce its own weights via weightwise application on itself (weight-wise self-application). In other words, fed with a single weight value and a corresponding positional encoding, the network learns a mapping to reliably reproduce each weight.
Since floating-point mathematics is imprecise in today's computers, we expect the weights to not be reproduced exactly but only within an $\varepsilon$ distance. Once a \acl{nn} can achieve sufficiently precise \acf{sr}, it is called an $\varepsilon$-fixpoint.

\begin{thm:def}[$\varepsilon$-fixpoint]\label{def:e-fixpoint}
	Given a neural network $\mathcal{N}$ with weights $\overline{\mathcal{N}} = \langle v_i \rangle_{0 \leq i < |\overline{\mathcal{N}}|}$. Let $\varepsilon \in \mathbb{R}$ be the error margin of the fixpoint property. Let $\mathcal{N}' = \mathcal{N} \appliedww \mathcal{N}$ be the self-application of $\mathcal{N}$ with weights $\overline{\mathcal{N}'} = \langle w_i \rangle_{0 \leq i < |\overline{\mathcal{N'}}|}$. We call $\mathcal{N}$ an $\varepsilon$-fixpoint or a fixpoint up to $\varepsilon$ iff for all $i$ it holds that $|w_i - v_i| < \varepsilon$.
\end{thm:def}

In most applications of artificial intelligence, the weights for neural networks are generated via training.

\begin{thm:def}[training]
Given a training data set $\mathbb{D} = \{(X_i, Y_i) \; | \; 0 \leq i < |\mathbb{D}|\}$ with data points consisting of an input $X_i$ and a desired output $Y_i$. We train a network $\mathcal{N}$ on $\mathbb{D}$ when we create a new network $\mathcal{N}'$ with adjusted weights $\overline{\mathcal{N'}}$ so to minimize the loss $\sum_{i=0}^{|\mathbb{D}|-1} |\mathcal{N'}(X_i) - Y_i|$ compared to $\mathcal{N}$. This update of weights is written as $\mathcal{N}' = \mathcal{N} \trained (X_0, Y_0) \trained ... \trained (X_{|\mathbb{D}|-1}, Y_{|\mathbb{D}|-1})$ or shorter $\mathcal{N}' = \mathcal{N} \trained \mathbb{D}$.
\end{thm:def}

Training of Feed-Forward Neural Networks (\emph{FF NN}), is usually performed via backpropagation (\emph{BP}, cf.~\cite{rumelhart1986learning}). 
When training self-replicating NNs by BP, we can define a weightwise training operator $\trainedww$ as follows:

\begin{thm:def}[weightwise training]
	Given \aclp{nn} $\mathcal{M}, \mathcal{N}, \mathcal{O}$ with $\overline{\mathcal{M}} = \langle v_i \rangle_{0 \leq i < |\mathcal{M}|}$. A neural network $\mathcal{O}$ is the result of weightwise training of $\mathcal{N}$ for $\mathcal{M}$, written $\mathcal{O} = \mathcal{N} \trained _\textit{ww} \mathcal{M}$, iff $$\mathcal{O} = \mathcal{N} \trained \{((v_i, L(i), C(i), E(i)), v_i) \; | \; 0 \leq i < |\overline{\mathcal{M}}|\}$$ where $L, C, E$ are defined as in Definition~\ref{def:appliedww}.
\end{thm:def}

As shown by Chang and Lipson~\cite{chang2018neural}, subjecting a randomly initialized neural network $\mathcal{N}$ to iterated self-application, i.e., computing $\mathcal{N} \appliedww \; \mathcal{N} \appliedww \; ... \appliedww \; \mathcal{N}$, causes the network to converge/diverge to trivial fixpoints $\overline{\mathcal{N}} \in \{\mathbf{0}, \mathbf{+\infty}, \mathbf{-\infty}\}$. Gabor et al.~\cite{gabor2019self} have shown that subjecting randomly initialized neural networks to self-training, i.e., computing $\mathcal{N} \trainedww \mathcal{N} \trainedww ... \trainedww \mathcal{N} \appliedww \; \mathcal{N}$, results in these networks becoming (usually non-trivial) $\varepsilon$-fixpoints up to small $\varepsilon \leq 10^{-7}$.

\subsection{Revisiting Additional Goals}

To formalize the \acf{at}, we include the findings of Gabor et al.~\cite{gabor2021goals} and adopt the procedure of their secondary (auxiliary) goal interface. 
They append auxiliary goals of the form $f : \mathbb{R}^2 \to \mathbb{R}$ to networks for weightwise application for a combined input format of $\mathcal{N} : \mathbb{R}^6 \to \mathbb{R}$. Since in our case, the auxiliary task will consist of multiplying one incoming scalar, we can simply omit the second auxiliary position and reuse this setting for our purpose. To reflect this change, we will update the wording of relevant auxiliary task definitions, briefly:

\begin{thm:def}[replicative and auxiliary application]\label{def:repauxapplied}
	Given neural networks $\mathcal{M}, \mathcal{N}, \mathcal{O}: \mathbb{R}^5 \to \mathbb{R}$ with $\overline{\mathcal{M}} = \langle v_i \rangle_{0 \leq i < |\mathcal{M}|}$.  Neural network $\mathcal{O}$ is the result of \emph{replicative application} of $\mathcal{N}$ for $\mathcal{M}$, written $\mathcal{O} = \mathcal{N} \appliedrep \mathcal{M}$, iff $$\overline{\mathcal{O}} = \langle \mathcal{N}(v_i, L(i), C(i), E(i), 0)\rangle_{0 \leq i < |\overline{\mathcal{M}}|}$$ where $L, C, E$ are defined as in Definition~\ref{def:appliedww}.
	
Given an input value $x \in \mathbb{R}$, the output value $y \in \mathbb{R}$ is the result of \emph{auxiliary application} of input $x$, written $y = \mathcal{N} \appliedaux X$, iff $$y = \mathcal{N}(0, 0, 0, 0, x).$$
\end{thm:def}

\section{Introducing the Organism Network}
\label{sec:Organism_Network_Intro}
Single networks of the structure described in the last subsection have proven to learn the ability to not only fulfil the \ac{sr}-property (cf. Definition~\ref{def:e-fixpoint}), but also complete an \acl{at}. 
In recent experiments, while \qquote{living} in the same environment (interacting from time to time through well-defined operators in the context of an artificial chemistry system), their \qquote{lives} were focused on either learning only their individual \ac{sr}-function \cite{gabor2022self_journal} or \ac{sr} plus an additional \acl{at} \cite{gabor2021goals}.
Those individual \acp{at}, even though being the same for all \ac{sr}-networks, are still only assigned to and learned by individual \acp{srnn} and do not construct any interactions (between the individuals) or incentive to work together.
In this work, we now introduce the concept to learn a shared task (\ac{at}), for which each \ac{srnn} plays its individual part.
Our end-to-end trainable architecture (in which the SRNNs are embedded) is built to fulfil a global task, while enabling \acp{srnn} to learn to self-replicate.
Inspired by the concept of biological organism, we define a hierarchical structure named \acf{on} that is built from groups (cells) of individual \acp{srnn}.
As we use a very specific interface for our implementation of the \ac{srnn}, we also refer to them as \acfp{pn}.

The requirements are as follows: 
    (1) The structure has to work with the given \ac{pn} interface so that the self-replication property can still be achieved.
    (2) The individual \aclp{pn} should still be able to interact with each other, hence a common task interface.
    (3) In addition to that, all \acp{pn}' \acl{at} functions must work within a defined structure so that the \acl{on} as a whole can fulfil a higher purpose.
    
\begin{figure}[tb]
    \begin{center}
        \includegraphics[width=\columnwidth]{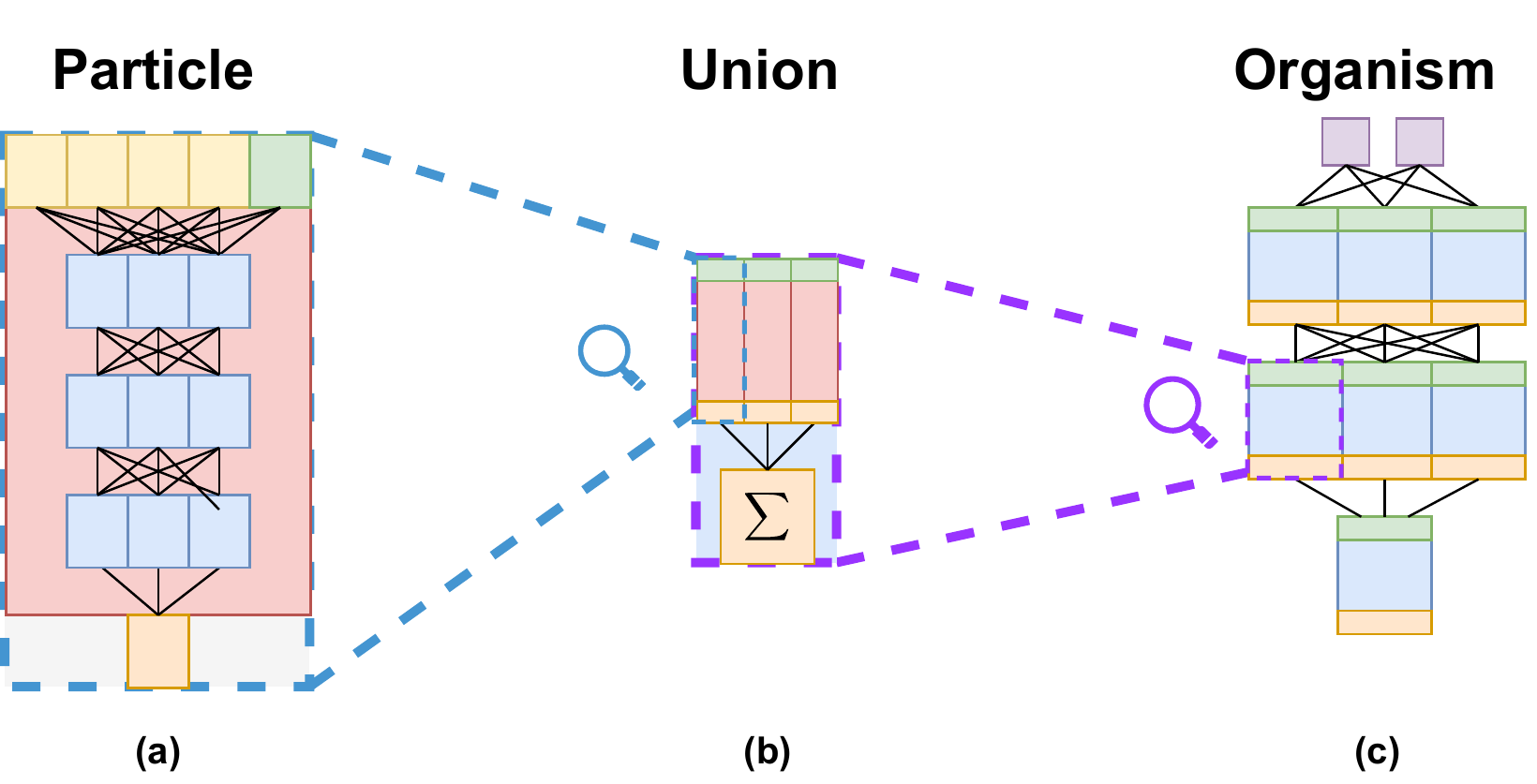}
        \caption{Schematic overview of the \ac{on} architecture. \textbf{(a)} A \ac{pn} approximates a weight application (output y, small orange box) for a given input value (x, green box), embedded in the auxiliary task input format of  Definition~\ref{def:repauxapplied} (yellow boxes). \textbf{(b)} A cell aggregates many predictions of the individual \acp{pn} composing them (orange bar/vector) into one sum per union (big orange box). \textbf{(cc)} The full \ac{on} (here: two inputs, three hidden layers) finally invokes many cells, bundled layer-wise as is common for deep NNs.
        }
        \label{fig:model_architecture}
    \end{center}
\end{figure}

To meet these requirements, we define the \ac{pn}s' \acf{at} (or goal) as the linear function of \qquote{multiply $x$ by a fixed value $w$}.
Figure~\ref{fig:model_architecture}a depicts the \ac{pn} and the described \ac{srnn} interface (grey, green).  
While only exposing the \acl{at} input (green) by passing \emph{zero} to the unused \ac{sa} inputs, we can define unions of \acp{pn} (Figure~\ref{fig:model_architecture}b) whose outputs are added up to form a single group response.
These unions which compute $\sum^0_i(PN_i(X_i))$ are commonly known as \aclp{c} in a \acl{nn}.
In other words, we propose a structured computational graph-architecture (\ac{on}, Figure~\ref{fig:model_architecture}c) in which interconnected groups (\ac{c}, Figure~\ref{fig:model_architecture}b) of individuals (\ac{pn}, Figure~\ref{fig:model_architecture}a) are combined, so that each \ac{pn} then behaves like a cell's weight in an NN.

\begin{thm:def}[organism network]\label{def:organism_network}
An \acf{on} is a \acf{nn} (cf. Definition~\ref{def:neural_network}) whose edges are not populated by scalar weights $E_{l,c,e}$ but instead by \aclp{pn} $\mathcal{M}_{l,c,e}$ (as defined by the \acp{pn}' weight vectors $\overline{\mathcal{M}_{l,c,e}}$), i.e., $\mathbf{w} = \overline{\mathcal{ON}} = \langle \mathcal{M}_{l,c,e} \rangle_{l=1,...,r, c = 1, ..., |L_l|, e = 1, ..., |L_{l-1}|}$.

A \ac{on}'s output provided an input $\mathbf{x} \in \mathbb{R}^p$ is given via $\mathbf{y} = \mathcal{ON}(\mathbf{x}) = \langle O'(r,c) \rangle_{c=1,...,|L_r|} \in \mathbb{R}^q$ where $$O'(l,c) = \begin{cases} x_c & \textit{ if } l = 0, \\ \sum_{i=1}^{|L_{l-1}|} \mathcal{M}_{l,c,i}(0,0,0,0,O'(l-1,c)) & \textit{ otherwise.}\end{cases}$$

\end{thm:def}

Given the already established research in the field of SRNNs, this architecture (\ac{on}, Figure \ref{fig:model_architecture}) can be trained end-to-end as an arbitrary linear function approximator, while each \ac{pn} still fulfils its self-replication property.



\begin{thm:def}[goal fulfilment]\label{def:goalfulfill}
An \acf{on} of \aclp{pn} $\mathcal{M}_1, ..., \mathcal{M}_{|\mathcal{ON}|}$\footnote{For this definition we ignore the internal structure of the \aclp{pn} in correspondence to the graph edges they are attached to.} has fulfilled its goal (up to a precision of $\varepsilon,\zeta$) iff
\begin{itemize}
    \item all \aclp{pn} $\mathcal{M}_1, ..., \mathcal{M}_{|\mathcal{ON}|}$ are $\varepsilon$-fixpoints (cf. Definition~\ref{def:e-fixpoint}) and
    \item for all specified test data $(x, y)$ the network's output $y' = \mathcal{ON}(x)$ fulfils $|y - y'| \leq \zeta$.
\end{itemize}
\end{thm:def}

Due to the given hierarchical order of \ac{on}, \ac{c}, and \ac{pn}, we call the optimization toward the collaborative \aclp{at} of \acp{pn} \emph{global} --- while both \acl{sr} and \qquote{acting-as-a-weight} are considered \emph{local} tasks.

\section{Experiment \#1 --- Float Addition}
\label{sec:exp_1_float_addition}

To provide a proof-of-concept of the \acp{pn} indeed being able to cooperate as individuals on a global \acl{at}, we start with a simple task: the addition of two floating point precision numbers. 
This task has been shown to be learnable in conjunction with \ac{sr} for a single \ac{srnn} by Gabor et al.~\cite{gabor2021goals}. 
To train towards the fulfilment of both the \textit{global} \ac{at} and the \textit{local} \ac{sr} task, we need to alternate the training schedule between the two tasks.\footnote{We also tried pre-training one or the other task first and then switching to the other goal (respectively), but this has been shown to heavily disturb the pre-trained weight configuration of the \ac{on}. 
We therefore train in alternating fashion.}

Using Stochastic Gradient Descent (SGD; $lr=0.004$, $momentum=0.9$, $epochs=50$) we train each \ac{pn} 25 steps of self-replication for each of the 20 batches in the primary (addition) task-dataset. Sufficient sizes for the \acp{pn} and \ac{on} (\wrt cells \& layers; (3,3), (3,2)) were determined empirically.

\subsection{Performance, Stability, and Robustness}

\begin{figure}[tb]
    \vspace{-10pt}
    \begin{center}
    \includegraphics[width=0.8\columnwidth]{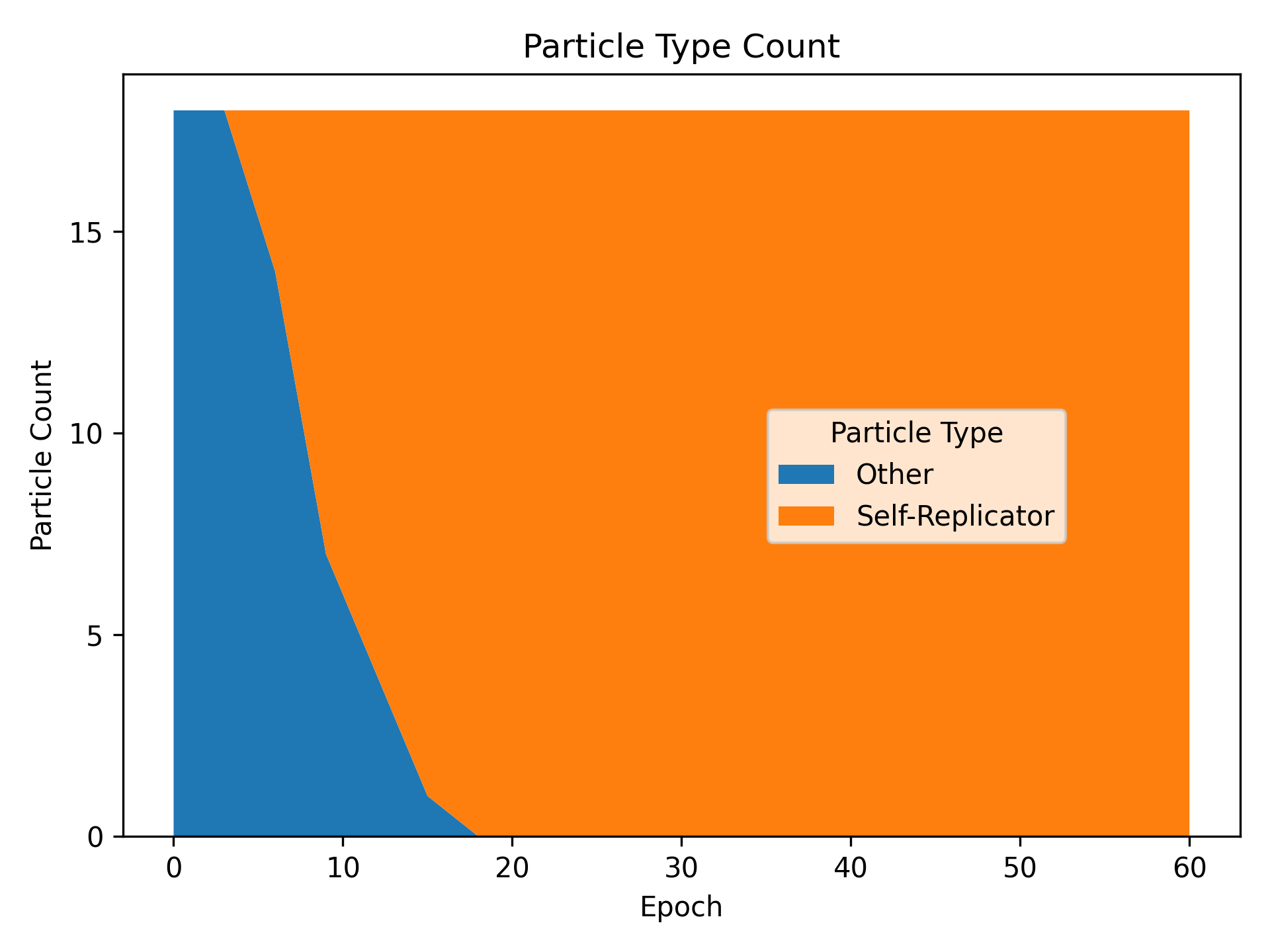}
    \caption{
        Distribution of \acp{pn} types (\pnsr~(orange), \pnf~(blue); \textbf{Y-axis}) composing the \ac{on} over the course of training experiment \#1, in epochs (\textbf{X-axis}). 
    }
    \label{fig:add-task-organism_type}
    \end{center}
    \vspace{-10pt}
\end{figure}

In accordance with Gabor et al.~\cite{gabor2019self,gabor2021goals} we tested the \acfp{pn} for \acf{sr} properties and studied their convergence and stability.
Figure~\ref{fig:add-task-organism_type} shows the timeline of all \acp{pn} within an \acl{on} slowly acquiring the \ac{sr}-capability over the course of training.

\begin{figure}[tb]
    \vspace{-10pt}
    \begin{center}
    \includegraphics[width=\columnwidth]{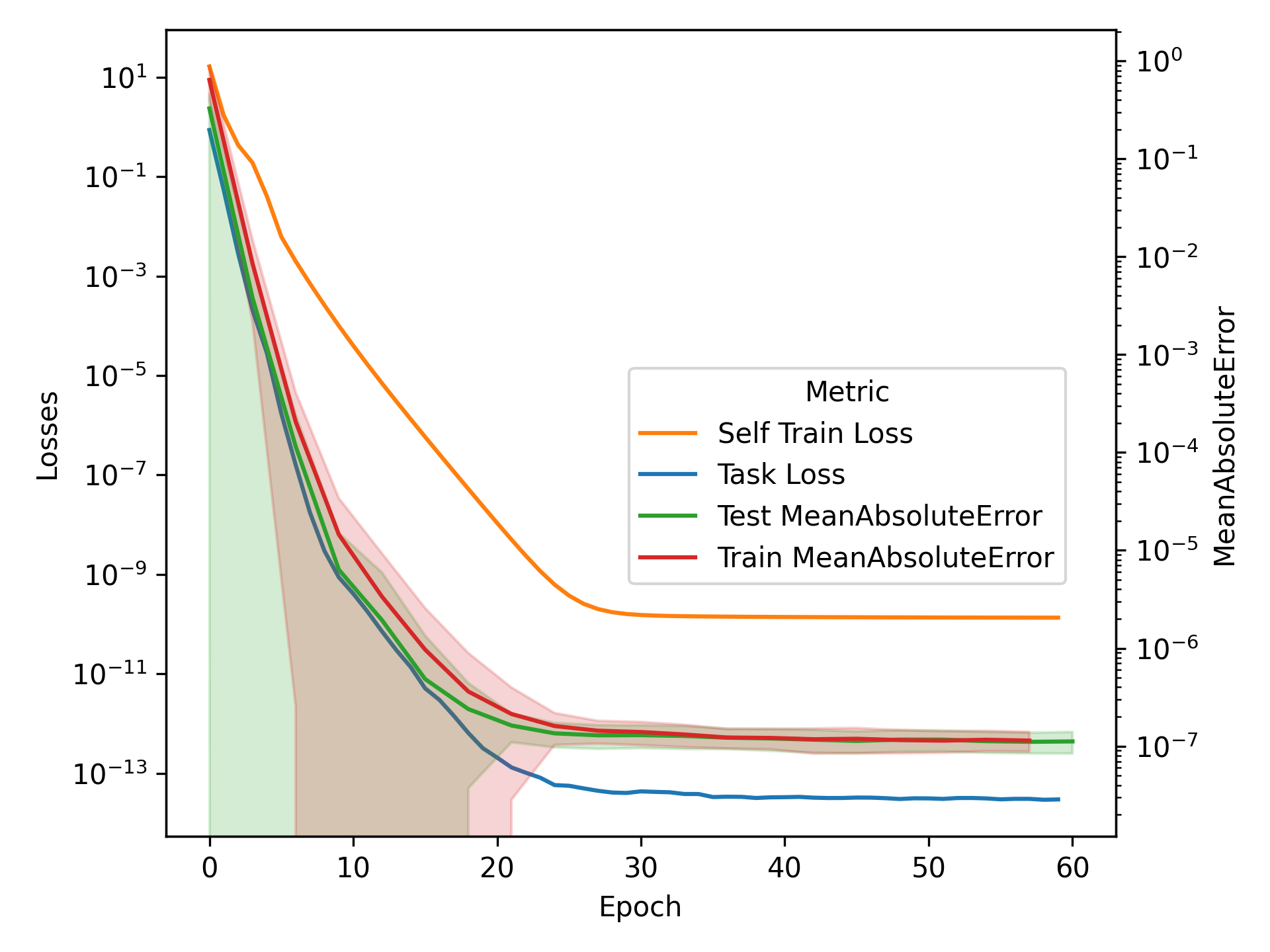}
    \caption{
        Mean squared error (MSE, \textbf{left Y-axis}) for both the \textit{local} \ac{sr}-task-loss (orange) and the \textit{global} add-task-loss (blue), vs. the mean absolute error (MAE, \textbf{right Y-axis}) of the \textit{global} add-task-loss, for comparison, over the course of training (red) and testing (green) ($n_{seeds}=10$, $epochs=60$, \textbf{X-axis}).
    }
    \label{fig:add-task-training_lineplot}
    \end{center}
    \vspace{-10pt}
\end{figure}

Even through the variance of several runs, we can observe both the local self-train loss and, shortly thereafter, the global addition task loss reaching their best predictions within the span of relative few epochs (\cf Figure~\ref{fig:add-task-training_lineplot}).
This speaks in favor of a robust and stable training behaviour, which is not surprising given the yet linear nature of the \ac{on} architecture.
With both averaged losses (\acl{sr} and \acl{at}) well below the $\varepsilon$ threshold, we consider all \acp{pn} as well as the over-spanning \ac{on} as goal-networks according to Definition~\ref{def:goalfulfill}.

\begin{figure}[htb]
    \begin{center}
    \includegraphics[width=0.95\columnwidth]{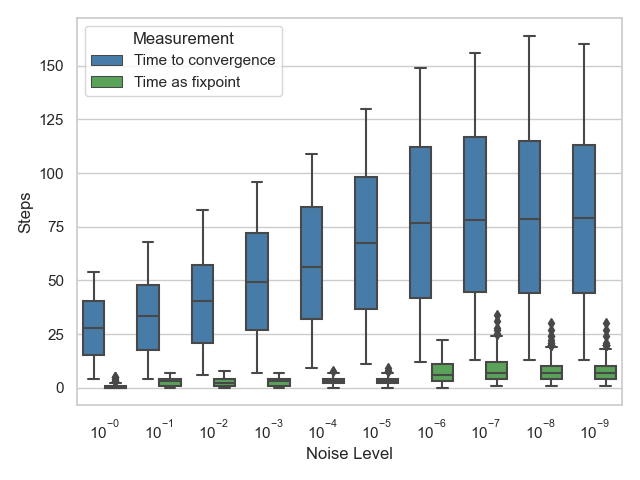}
    \caption{
      The robustness of \acp{pn} that did learn the \ac{sr}-property \pnsr \wrt the weight-wise application $\lhd_\textit{ww}$ (\cf Definition~\ref{def:appliedww}). 
      \textbf{X-{axis}:} Range of noise applied on the initial input. 
      \textbf{Y-{axis}:} Steps of self-application \acp{pn} keep their \ac{sr}-property \wrt $\varepsilon=10^{-5}$ (\textbf{green}); 
      steps of self-application until the network is regarded as diverged (\textbf{blue}).}
    \label{fig:add-task-robustness}
    \end{center}
    \vspace{-10pt}
\end{figure}

Regarding the robustness of the \ac{sr} property, we borrow the repeated \acl{sa} experiment conducted by Gabor et al.~\cite{gabor2022self_journal} for our \acl{on} (\cf Figure \ref{def:appliedww}). 
We found that, in comparison to both other studies, the \acp{pn} hold their \ac{sr} property for longer when applied to themselves (\cf Definition~\ref{def:appliedww}), some even under the influence of strong Gaussian noise ($>10^{-5}$) on the initial inputs (\cf Figure \ref{fig:add-task-robustness}).
Interestingly, even though the observed variance as well as the $.95$ confidence interval span over a wide margin, the average robustness per noise level is higher than shown for simple self-replicators, even perfectly constructed ones (\cf Gabor et al.~\cite{gabor2022self_journal}, Figure 3).
Therefore, \pnsr seem to be even stronger $\varepsilon$-fixpoint generators than individually trained \acp{srnn}.

\subsection{Weight Re-Substitution}

\begin{figure}[htb]
    \vspace{-10pt}
    \begin{center}
        \includegraphics[width=0.9\columnwidth]{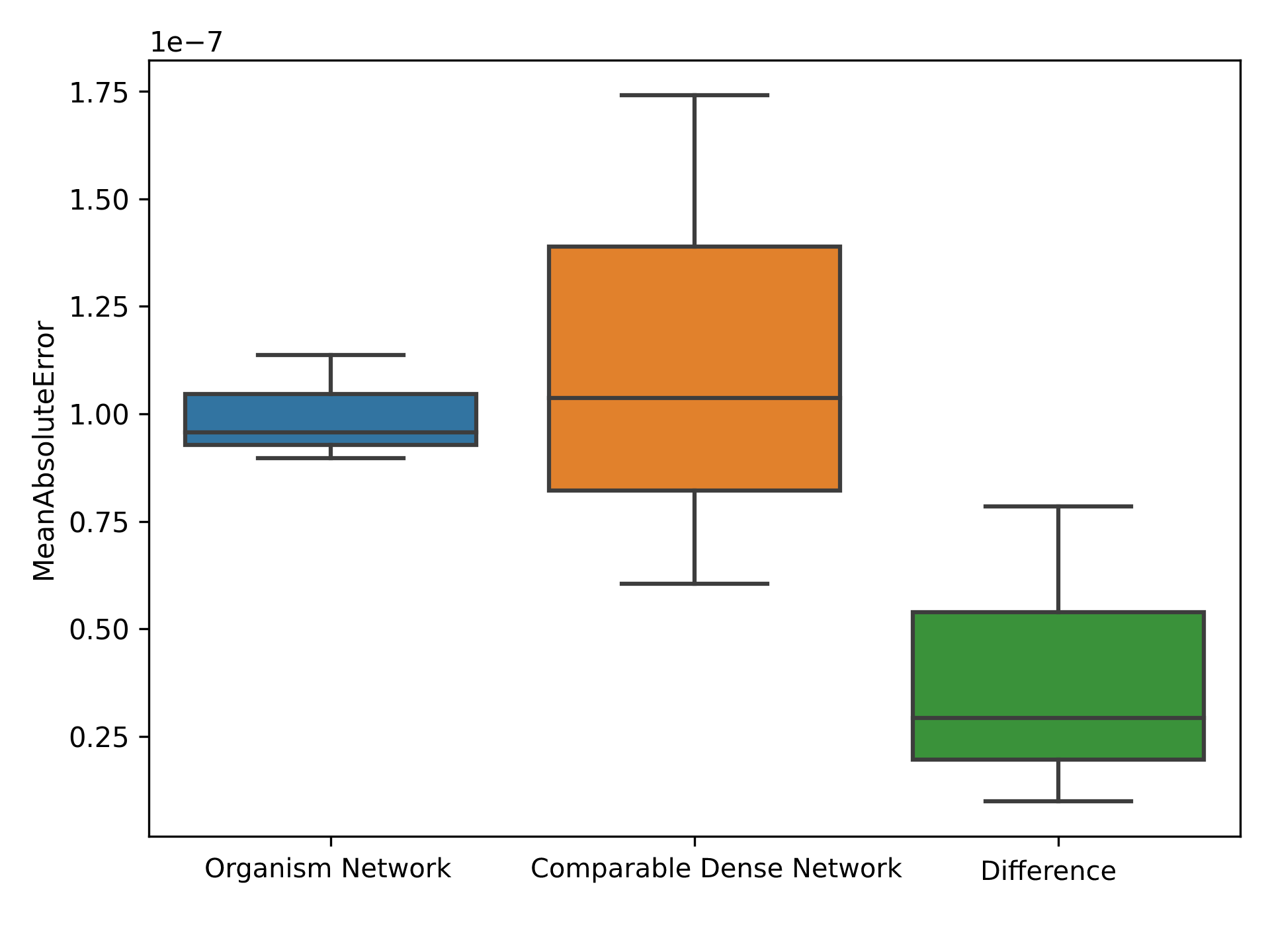}
        \caption{Absolute margin of error for the addition-task (\textbf{Y-axis}, scaled at 1e-7; Definition~\ref{def:repauxapplied}) after replacing parameters in a feed-forward \ac{nn} of same size (w.r.t number of neurons and layers) by \textit{learned weight application} of $t \in PN_0^i$ (\cf Def.~\ref{def:dropout_test})
        }
        \label{fig:sanity_margin_task}
    \end{center}
    \vspace{-10pt}
\end{figure}

To check whether the trained \acfp{pn} are behaving like weights in an \ac{nn}, we introduce a fixed input of $x_{ON_w} = (0,0,0,0,1)$, matching the organism interface. 
The \acp{pn}' output value, as the product of $1$ times the weight-scalar that the organism learned to represent (in a linear network), represents the extracted weight value.
We then replace a conventional feed-forward \ac{nn} of same size (\wrt number of neurons and layers) of \ac{on} parameters with the extracted weight values and calculate the absolute margin of error for the classification task. 
This experiment shows that our \ac{on} of approximated weight applications behaves like regular neural networks (by a margin of error of $\leq 10^{-8}$) as shown in Figure~\ref{fig:sanity_margin_task}, but at a noticeable lower variance.

\section{Experiment \#2 --- MNIST Image Classification}  
\label{sec:exp_2_mnist_classification}

After confirming the basic validity of the organism network on the \qquote{float-addition} task we now explore the possibility of training a bigger network with image classification as the global task. 
In this experiment, we intend to move the context of self-replicating organisms (and \acp{srnn}) closer to real-world applications. 
As the training time and compute costs grow quickly with additional input dimensions and each additional \ac{pn}, we decide in favour of a downsized variant for the scope of this research. 
While there may be more optimized setups in the future, we used a $(15 \times 15)$ pixel variant of \ac{mnist} (cf. Deng~\cite{deng2012mnist}).
This dataset consisting of $70,000$ images of standardized handwritten numbers ($30 \times 30$px) was considered a standard deep-learning benchmark for several years.
As \ac{mnist} is much harder to learn than simple addition, we prioritize the image classification task by adjusting the self-train vs. \textit{global}-task-train schedule to a 5:1 ratio. 
To enable better organism network training, we also increase the model width and the hidden layer depth to 3 (excluding the input and output layers). 
Also, we introduce non-linearity in the form of the \ac{gelu}~\cite{hendrycks2016gaussian_gelu} activation function after every hidden layer.
The optimizer setup remains with the same parameters.

\subsection{Performance, Stability, and Robustness}
\begin{figure}[tb]
    \begin{center}
    \includegraphics[width=\columnwidth]{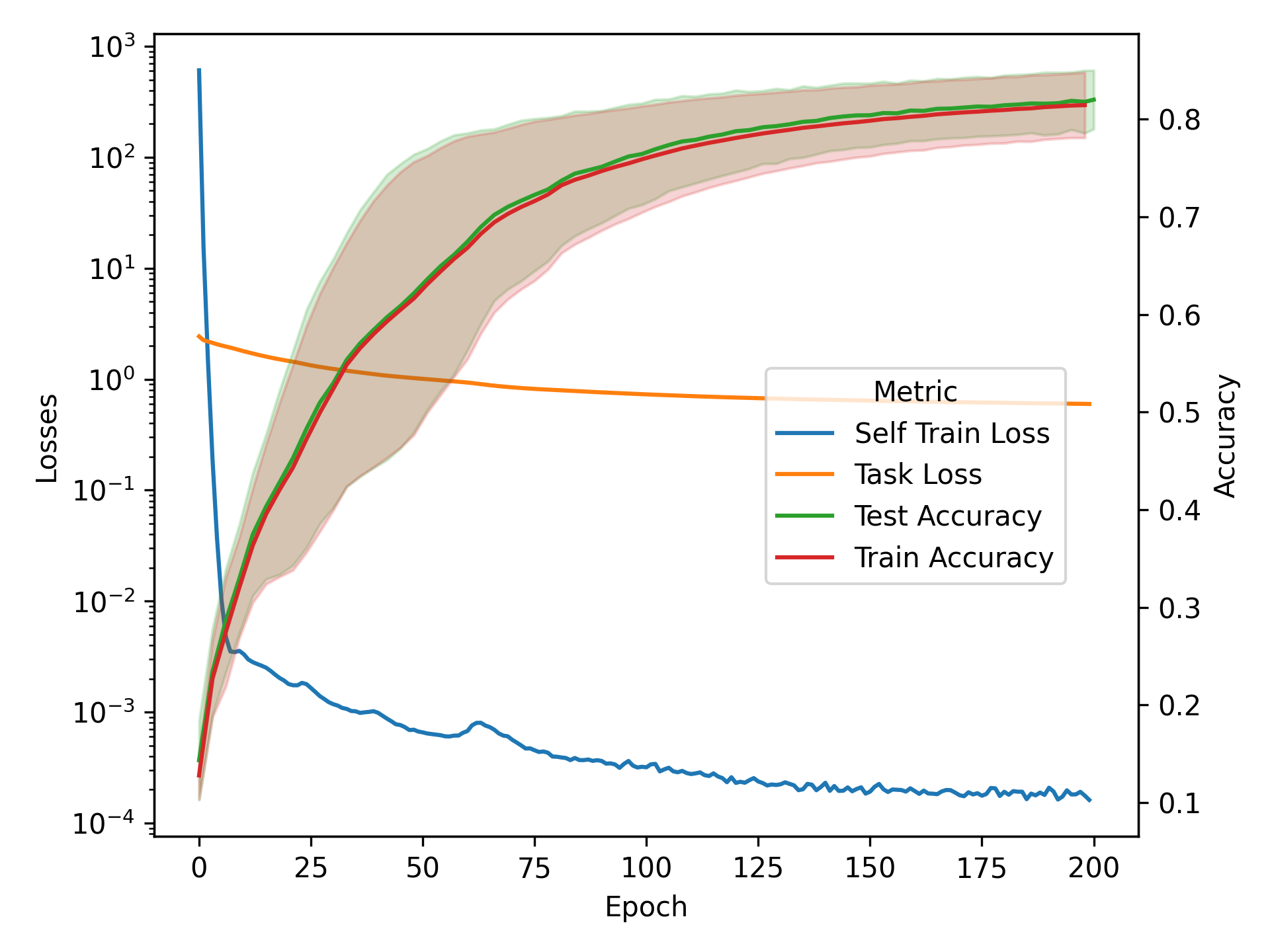}
    \caption{
        Losses (\textbf{left X-axis}) vs. accuracy (\textbf{right Y-axis}) for both, the \textit{local} SR-task (loss blue) and the \textit{global} \emph{MNIST}-classification-task (loss orange) over the course of training (red) and testing (green) ($n_{seeds}=3$, $epochs=200$, \textbf{X-axis}).
        }
    \label{fig:mnist-training_lineplot}
    \end{center}
    \vspace{-10pt}
\end{figure}

Figure~\ref{fig:mnist-training_lineplot} shows the losses of both tasks and the accuracies. 
Testing was carried out on a withheld 10,000 sample test set, as is convention (see Deng~\cite{deng2012mnist}).
We observe the \ac{sr}-loss improving much faster than expected, with the classification accuracy growing steadily to convergence. 
Although the results may not be state-of-the-art precision, we deem a test accuracy of $>0.8$ on an (in effect) tiny network with no bias a decent example of the \ac{on}'s capability.

\begin{figure}[htb]
    \vspace{-10pt}
    \begin{center}
    \includegraphics[width=0.95\columnwidth]{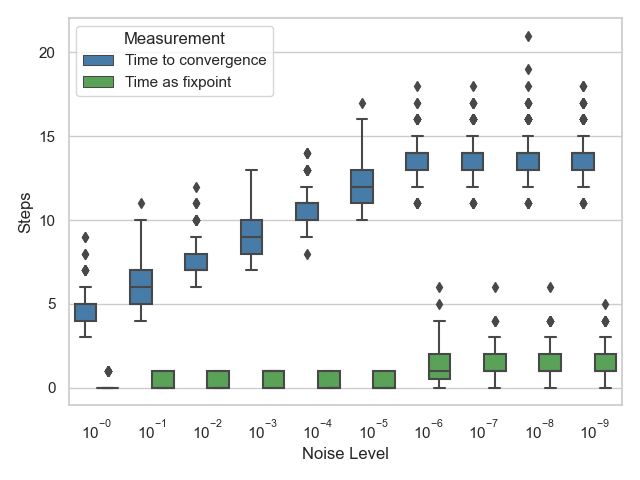}
    \caption{
      The robustness of \acp{pn} that did learn the \ac{sr}-property \pnsr regarding the weight-wise application $\lhd_\textit{ww}$ (\cf Definition~\ref{def:appliedww}). 
      \textbf{X-{axis}:} Range of noise applied on the initial input. 
      \textbf{Y-{axis}:} Steps of self-application \acp{pn} keep their \ac{sr} property with $\varepsilon=10^{-5}$ (\textbf{green}); 
      Steps of self-application the network was regarded as diverged (\textbf{blue}).
    }
    \label{fig:mnist-task-robustness}
    \end{center}
    \vspace{-10pt}
\end{figure}

When revisiting the test for robustness (\cf Figure~\ref{fig:mnist-task-robustness}) on \acp{pn} trained on \ac{mnist}, we observe much lower levels of robustness, which are comparable to findings in the established literature.
This contrasts our previous results and leads to the assumption that, with longer training and regimes which are more focused on the \ac{sr}-task, very robust and stable self-replicators are achievable.

After an in-depth hyperparameter search, we observed a major drawback: Even though the \ac{sr}-loss converged fast, Figure~\ref{fig:mnist-organism_type} shows that just about a quarter of all \acp{pn} gained the \ac{sr}-property.
This contrasts with our original assumption because we had expected that all \ac{pn} would achieve this property.

\subsection{Specialist Dropout Test}

As it is not possible for us to train all \acp{pn} to learn both tasks at the same time, we decide to further investigate the abilities of the \acp{pn}.
We assume that training on \ac{mnist} is more complex than our simple Experiment \#1; \ac{pn} (weights) with less importance for the \acl{at} gain the \ac{sr}-property~\pnsr earlier than the others~\pnf.
To validate our assumption, we construct a dropout test comparable to a \qquote{per-weight-dropout} (i.e., making an \ac{nn} sparse by pruning), in which we test the influence of both groups $[PN_{SR}, PN_F]$ on the task accuracy (\cf Definition~\ref{def:dropout_test}). 
In other words, we want to know how important a subset of weight positions is (defined by its \ac{sr}-property) for solving the overall group task. 

\begin{thm:def}[dropout organism network]\label{def:dropout_test}
Let $\mathcal{ON}$ be an \acl{on}, so that $\overline{\mathcal{ON}} = \langle \mathcal{PN}_i \rangle_{i=1,...,|\overline{\mathcal{ON}}|}$ is a vector of particle networks $\mathcal{PN}_i : \mathbb{R}^5 \to \mathbb{R}$ for $i=1,..., |\overline{\mathcal{ON}}|$.
Let $T : (\mathbb{R}^5 \to \mathbb{R}) \to \{PN_{SR},PN_F\}$ be a labelling of all particle networks, with the implication that any self-replicating particle network $\mathcal{PN}$ is labelled with $T(\mathcal{PN}) = \mathcal{PN}_{SR}$ and any other particle network $\mathcal{PN}$ is labelled with $T(\mathcal{PN}) = PN_F$.
Let $\mathcal{Z} : \mathbb{R}^5 \to \mathbb{R}$ be the zero network so that $\mathcal{Z}(\_,\_,\_,\_,\_) = 0$ for any input. The dropout organism network $\mathcal{ON}'$ for $\mathcal{ON}$ w.r.t. type $t \in \{PN_{SR},PN_{F}\}$ is given via $\overline{\mathcal{ON}} = \langle D(i) \rangle_{i=1,...,|\overline{\mathcal{ON}}|}$ where $D(i) = \begin{cases} \mathcal{Z} & \textit{if } T(\mathcal{PN}_i) = t,\\ \mathcal{PN}_i & \textit{otherwise.} \end{cases}$
\end{thm:def}

\begin{figure}[tb]
    \begin{center}
        \includegraphics[width=0.8\columnwidth]{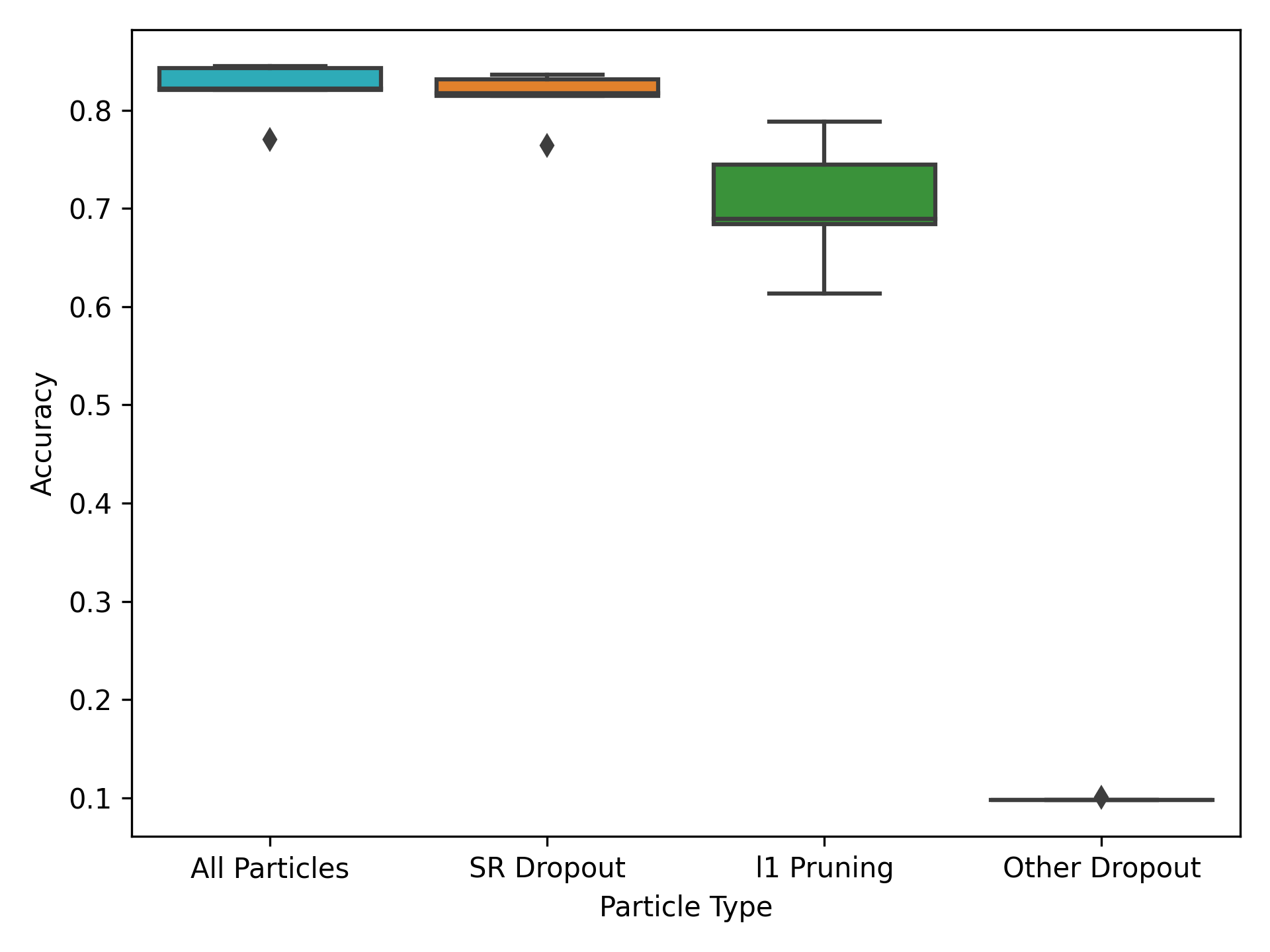}
        \caption{
            Accuracy (\textbf{Y-axis}) after per-weight-dropout for types $t \in \{PN_{SR}, PN_F\}$ compared to whole organism and `l1-norm' pruning (\textbf{X-axis}).
        }
        \label{fig:mnist-dropout_test}
    \end{center}
    \vspace{-10pt}
\end{figure}
\begin{figure}[tb]
    \begin{center}
    \includegraphics[width=0.8\columnwidth]{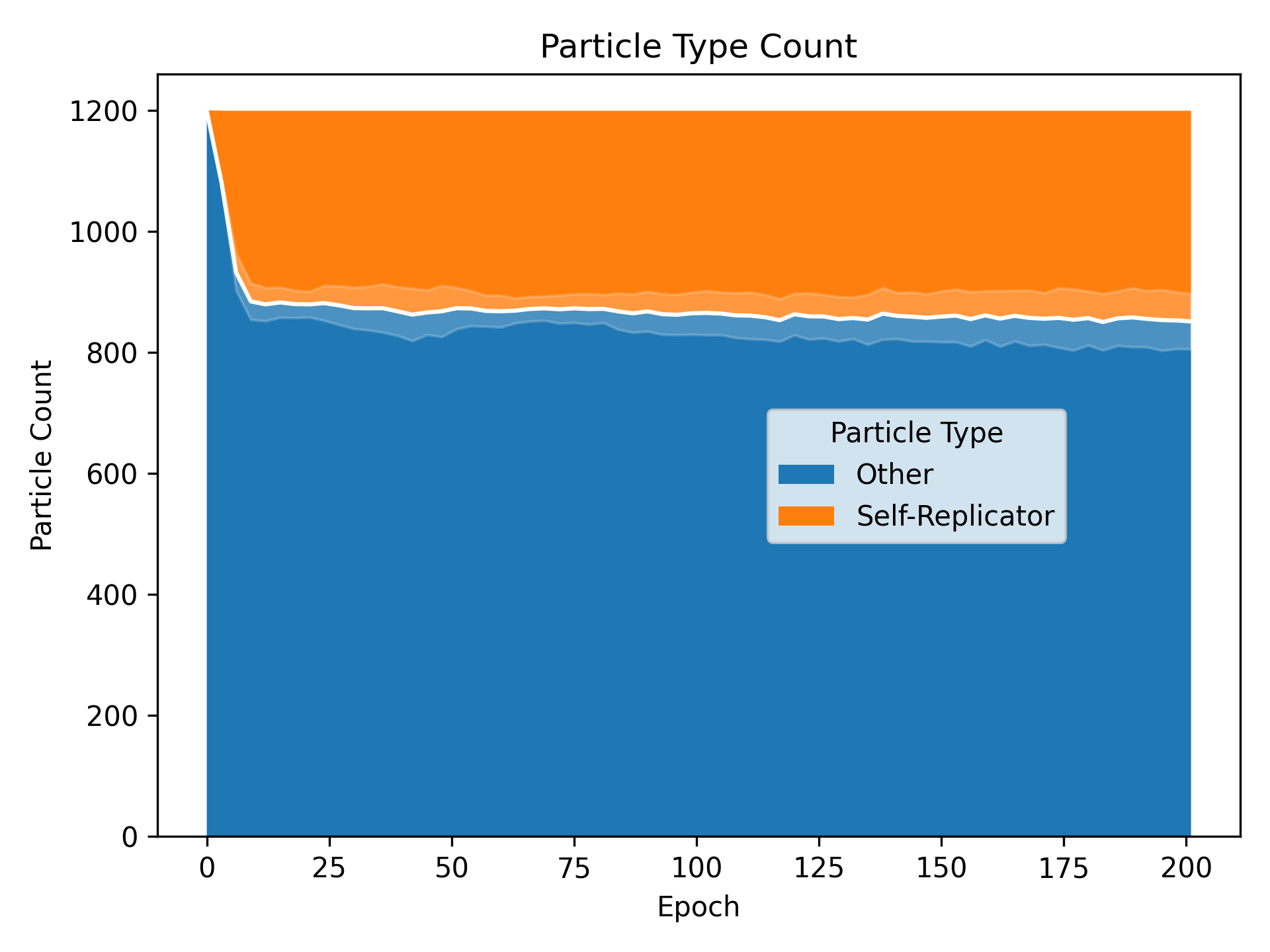}
    \caption{
        Distribution of \ac{pn} types (\pnsr (orange), \pnf (blue); \textbf{Y-axis}) over the course of training the \ac{on} on experiment \#2, in epochs (\textbf{X-axis}).
        We observe an initial saturation in the amount of \pnsr, which then remains roughly stable over the rest of training. We plot the variance over $n=5$ runs with a lighter color to show the consistency of this specialisation development.
    }
    \label{fig:mnist-organism_type}
    \end{center}
    \vspace{-10pt}
\end{figure}

Figure~\ref{fig:mnist-dropout_test} depicts the accuracy after per-weight-dropout for all groups of $t \in \{PN_{SR}, PN_F\}$ compared to the whole \acl{on}.
We observe that \pnsr-dropout results in almost no change in the measured test accuracy ($>0.001$ accuracy), which is quite surprising considering that we disable a sizable chunk of the network parameters.
The other way around, $PN_F$-dropout disables the \ac{mnist} classification immediately. 
This is the expected outcome, given the vast proportion of $PN_F$ in the $ON$ and their connectivity (\cf Figure~\ref{fig:mnist-connectivity}).
In other words, \acp{pn} which are able to \qquote{specialise} in self-replication are obviously not crucial for the group task, which allows us to reduce the networks' parameter size by approx. $25\%$.
However, we observe a slow but gradual trend towards more self-replicators (\pnsr) (\cf Figure~\ref{fig:mnist-organism_type}), suggesting insufficient training time so that clear convergence may have simply yet to occur, which, we believe, is not as likely, given the alternating training.
Instead of blindly training more epochs, we settle with this situation and analyse it further.

\begin{figure}[tb!]
    \begin{center}
        \includegraphics[width=\columnwidth, trim={0 3.7cm 0 3.7cm},clip]{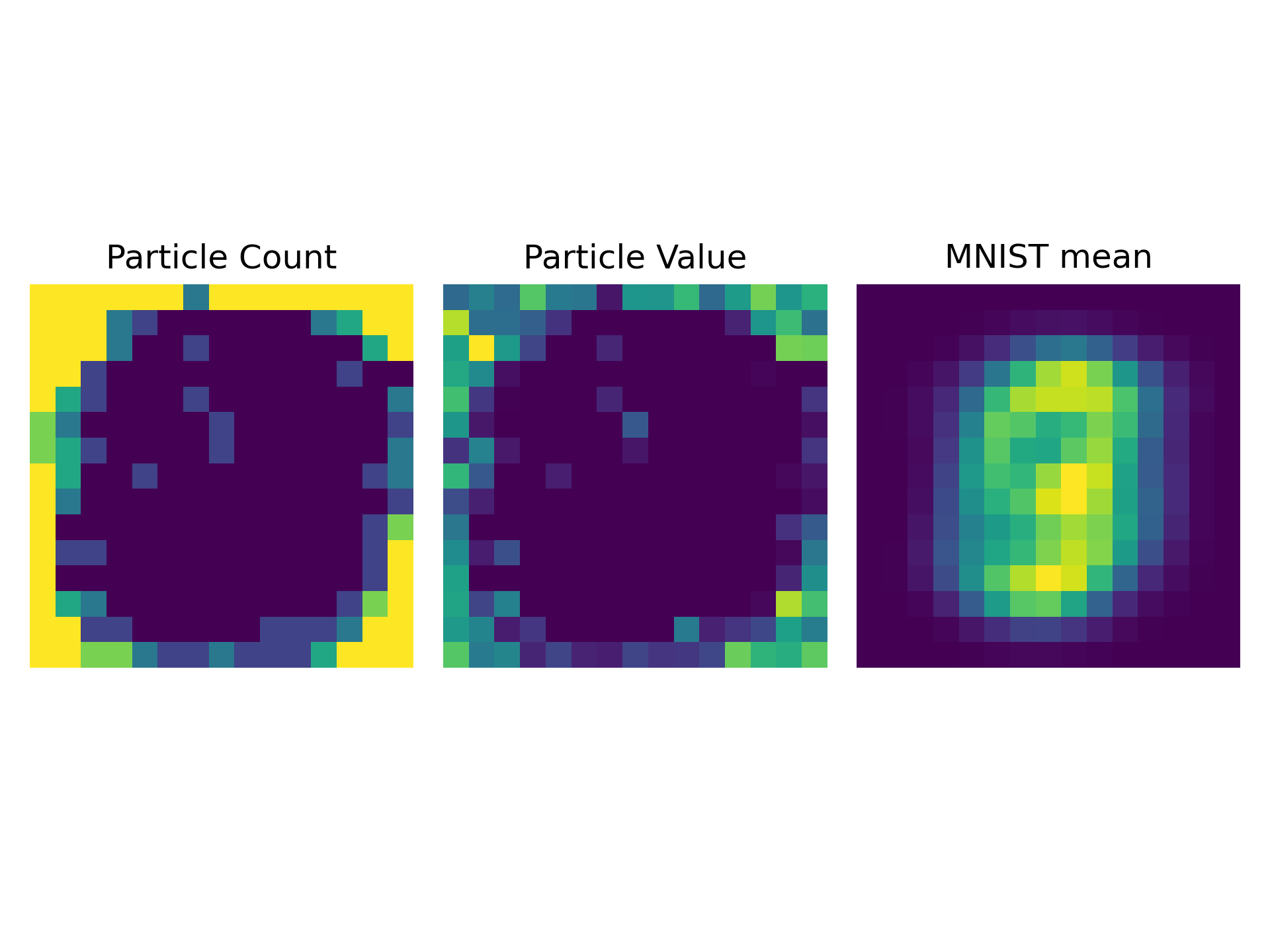}~\\
        ~\hspace{3.5em} (a) \hfill (b) \hfill (c) \hspace{3.5em}~
        \caption{
            Heatmap representing the position of found \pnsr. \textbf{(a)} Per pixel count of \pnsr over all $PN$ in the first $ON$-layer. Lighter colour represents more and darker colour fewer \pnsr responsible for this input. \textbf{(b)} Per-pixel-sum of the \qquote{equal-weight-value} of \pnsr overall $PN$ in the first $ON$-layer. \textbf{(c)} Per-pixel-mean of the \emph{MNIST} dataset for comparison. Here, lighter colour signifies higher values.
        }
        \label{fig:mnist-self-replicator-position}
    \end{center}
    \vspace{-10pt}
\end{figure}

\begin{figure}[tb]
     \centering
     \begin{minipage}[b]{0.45\columnwidth}
         \centering
         \includegraphics[width=\columnwidth]{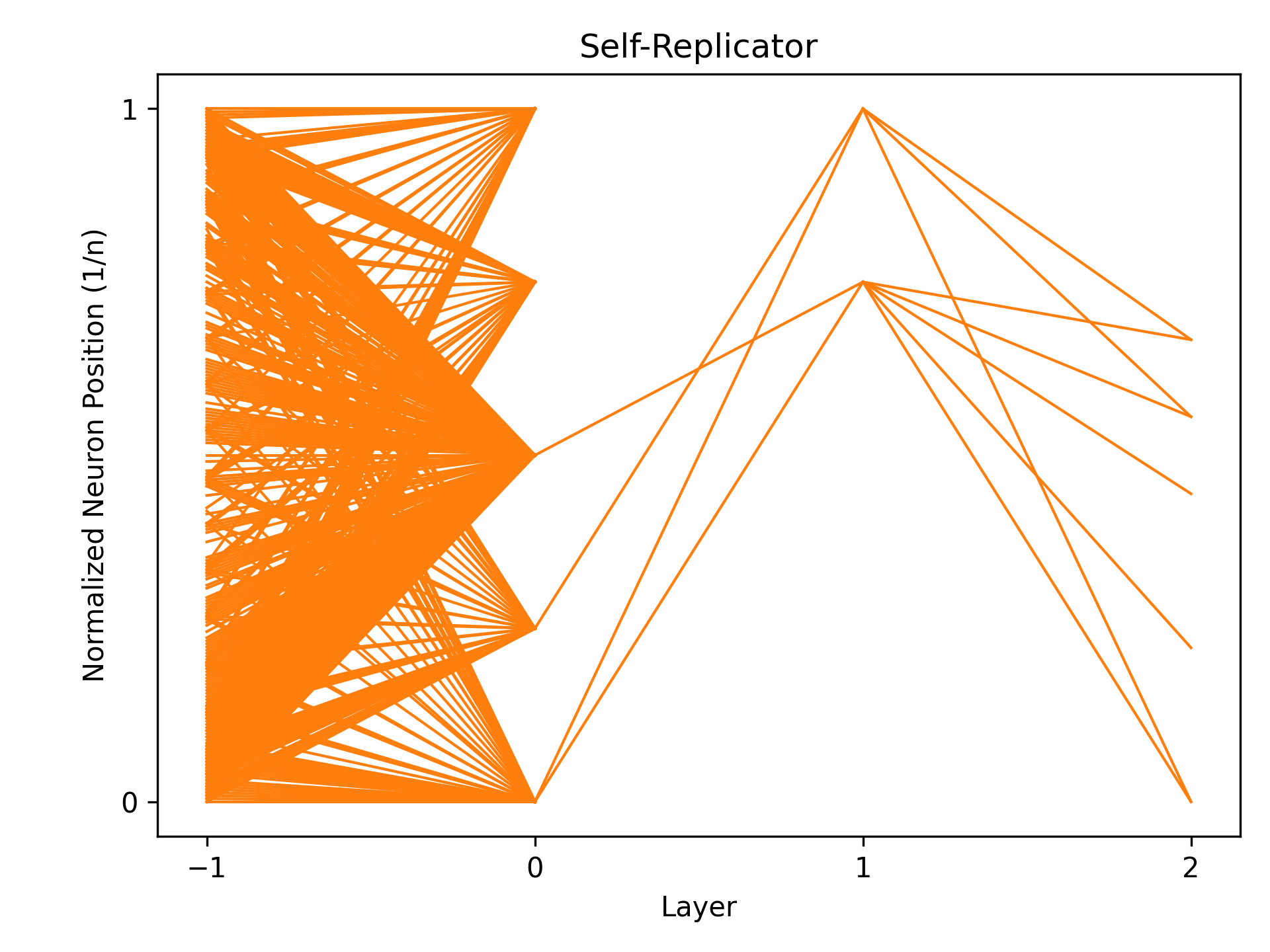}\\%
         (a)
     \end{minipage}
     ~
     \begin{minipage}[b]{0.45\columnwidth}
         \centering

         \includegraphics[width=\columnwidth]{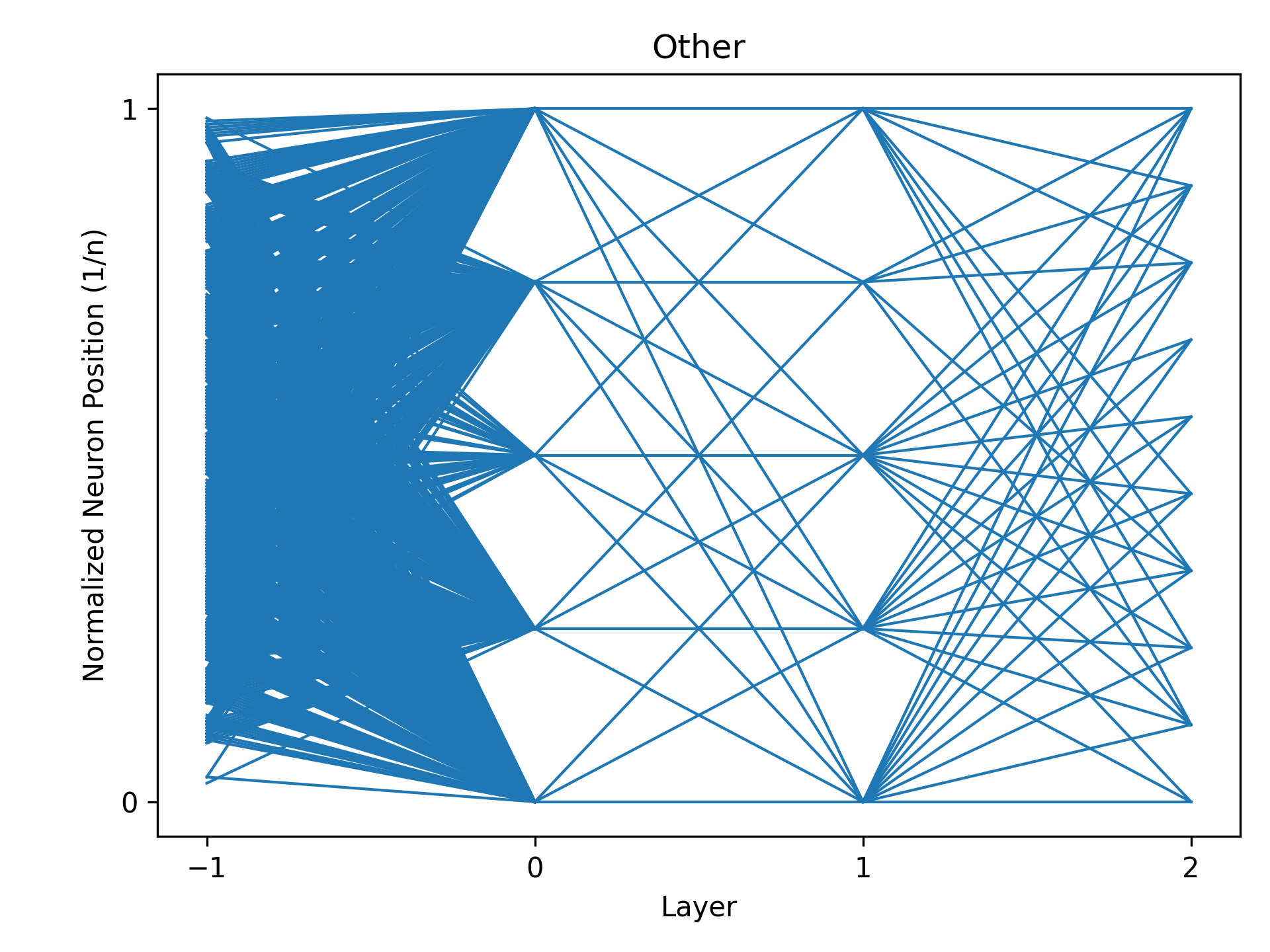}\\%
         (b)
     \end{minipage}
     \caption{Connectivity per organism group: 
     \textbf{(a)} \Acp{pn} that have learned the self-replication trait (\pnsr, orange); 
     \textbf{(b)} \Acp{pn} that have learned the global \acl{at} (\pnf, blue). (Layer \#-1 = \ac{on} input vector.)
     }
     \label{fig:mnist-connectivity}
    \vspace{-10pt}
\end{figure}

Since \ac{mnist} allows for a visual exploration, we visualise the positioning of \aclp{sr} with \ac{sr} property.
Figure~\ref{fig:mnist-self-replicator-position}a represents the position of \pnsr as \qquote{heatmap}, while Figure~\ref{fig:mnist-self-replicator-position}b is a sum of the approx. weight values (both \wrt $x \in X$ each PN is attached to).
We find it quite interesting that even the \pnsr which are not relevant for the task keep on developing weight representations. Those seem to then get ignored somewhere down the line.
This supports our assumption, that \pnsr for themselves, \qquote{realise} (presumably due to a small gradient) that they are assigned to \qquote{non-important} input vectors.
For comparison, Figure~\ref{fig:mnist-self-replicator-position}c visualises the per-pixel-mean of the test dataset.
We clearly see that replicator specialists (\pnsr) are mostly responsible for weight predictions of the empty parts of the image around the edges.
As \emph{MNIST} comes with a lot of \emph{zero}-values, we added \emph{Gaussian} noise of magnitude $10^{-4}$ to each image. 
This ensures that the observed behaviour is not simply the result of \acp{pn} not being used at all for the \ac{at}.

Figure~\ref{fig:mnist-connectivity} visualises the network connectivity per \ac{pn}-type. 
This reveals a heavy distribution of \pnsr in the input-layer, but not solely.  
We assume that this positioning is the direct result of \qquote{non-importance} in input pixels. leading to much faster convergence for \acp{pn} responsible for the image-edge areas. 
This is quite interesting, as we tried to counter the obvious \qquote{non-importance} by adding Gaussian noise, as previously described.

\begin{figure}[tb]
     \centering
     \begin{minipage}[b]{0.45\columnwidth}
         \centering
         \includegraphics[width=\columnwidth]{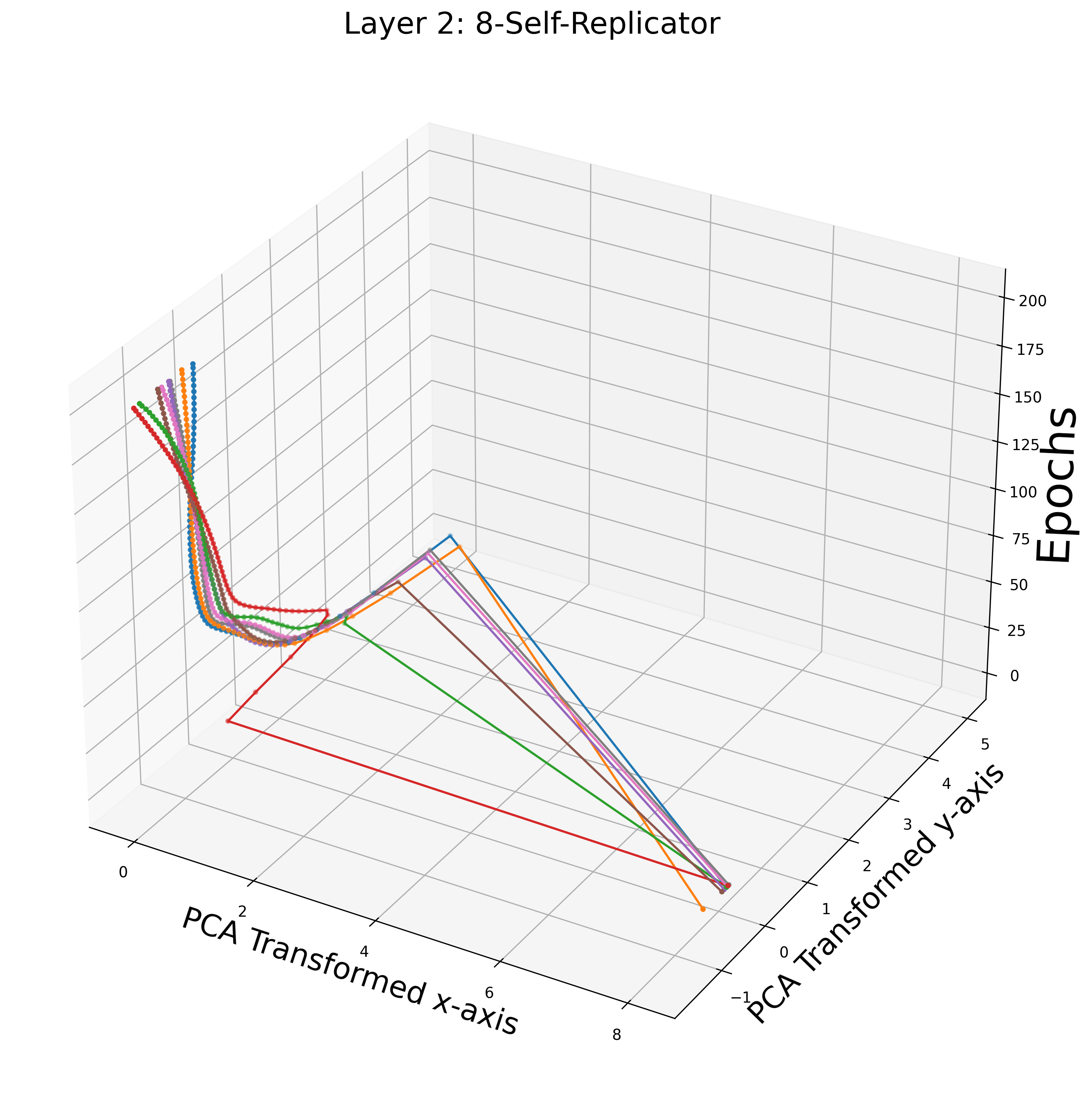}\\%
         (a)
     \end{minipage}
     ~
     \begin{minipage}[b]{0.45\columnwidth}
         \centering

         \includegraphics[width=\columnwidth]{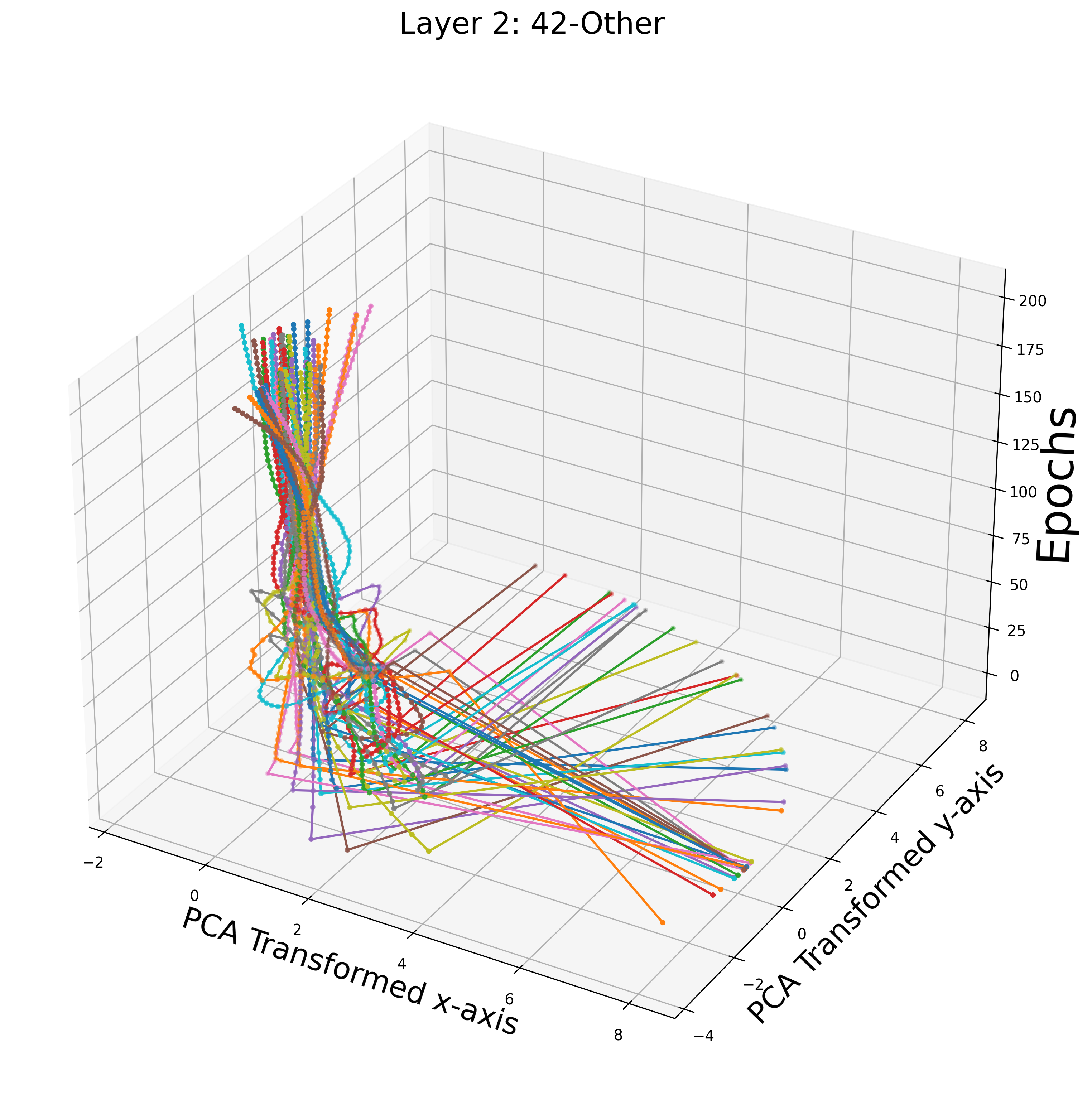}\\%
         (b)
     \end{minipage}
     \caption{
         PCA transformed weight-space trajectories (per \ac{pn}-group); here Layer 2 of the \ac{on}:
         \textbf{(a)} 8 \pnsr that have gained the \acl{sr}-trait; 
         \textbf{(b)} 42 \pnf that have learned the \acl{at}. Every trajectory (unique colour), shows the change in parameter-values of its corresponding \ac{pn}.
     }
     \label{fig:3d-trajectory-layer-2}
     \vspace{-4pt}
\end{figure}

In the style of Gabor et al.~\cite{gabor2022self_journal, gabor2021goals} we also take a look at the PCA-transformed weight trajectories per \ac{pn}-group of an interesting model (\cf Figure~\ref{fig:3d-trajectory-layer-2}).
Along most of the networks layers (we evaluated more than a single \ac{on}), we observe that \pnf weight trajectories quickly reach the specific region of their PCA weight space.
This is the expected behaviour.
Trajectories of \pnsr, in contrast, are mostly fitted along a single axis by PCA (shared PCA space) (\cf Figure~\ref{fig:3d-trajectory-layer-2}a; $layer=2$). 
This behaviour can also be observed with \pnsr in other networks/layers. 
Please note that we check for trivial fixpoints (e.g., when all weights become \emph{zero}).
Like Gabor et al.~\cite{gabor2019self}, we do not consider those \qquote{self-replicators} in general but \emph{zero-fixpoints}.
Due to over-plotting, we unfortunately did not learn a lot from PCA trajectories of the heavily populated first layer ($15 \times 15 \times 5$ \acp{pn}; cf. Figure~\ref{fig:mnist-connectivity}).

\section{Related Work}
\label{sec:related_work}

Self-replication as a field of research in the context of \aclp{nn} receives increasing attention in the past few years. 
Chang and Lipson~\cite{chang2018neural} developed self-replicating \emph{NN quine}, which they reviewed on the full \textit{MNIST} classification task.
In contrast to our findings, \emph{NN quines} seem to lose their \qquote{incentive} to receive or hold their \ac{sr} ability. 
Our approach in contrast follows the path of optimization to both (\emph{local} and \emph{global}) intermixed tasks.

Randazzoo et al.~\cite{randazzo2021fertile_self_replication} also established a procedure in which they (through \ac{bp}) create initial conditions that generate \qquote{fertile} self-replicating agents.
Training their SR property is achieved by introducing a \qquote{sink-loss} which targets a noisy neighbour's child (network configuration after \ac{sa}) rather than the network's own weights, thereby encouraging diverse network replication.

Embedding the replication task into larger structures has -- to the best of our knowledge -- not been done before. 
However, the idea of replacing neural nodes with small networks was proposed by CCamp et al.~\cite{camp2020continual_deep_neurons}, but they focus on the topic of continual learning. \
Their \acp{dan} share one meta-pretrained parameter set that minimizes a memory loss (of previously seen linear regression tasks). 
Keeping these pretrained parameters fixed, they try to learn (via \ac{bp}) a weight matrix, connecting the \acp{dan}.
The focus of this approach is set in favour of the malleability of these \qquote{vectorized synapses} (i.e., synaptic plasticity), rather than on the neurons. 
This two-weight-set approach is part of recent advances in continual learning research, reserving different weight partitions for task memory: 
Ba et al.~\cite{ba2016using}, e.g., present the benefits of using 2 synaptic weight sets with novel \qquote{fast weights} for improved knowledge retention. 
More recently, Hurtado et al.~\cite{hurtado2021optimizing} proposed specific shared weights as a common knowledge base between tasks. 
Our work, in contrast, does not check for the memory of tasks or explicitly assign task positions, but rather aims to train dual-purpose particles right from the start. 
We also do not keep any part of the organism fixed, but we do second the choice of utilizing small neuron particles, especially since Beniaguev et al.~\cite{beniaguev2021single_bio} have shown that (even small) multi-layer networks suffice in mimicking the functionality of real cortical neurons.

The process of removing the \pnsr from the trained \ac{on} (\cf our post-training dropout test) is comparable to the concept of selecting important network weights in the context of \ac{nn} pruning.
For a broad overview, please refer to Blalock et al.~\cite{blalock2020state}. 
This is of interest as, on the one hand, modern network architectures are assumed to be prone to over-parameterization \cite{denil2013predicting}.
However, on the other hand, recent research suggests that there are subsets of parameter configurations with (at least) equal test accuracy (compared to the original network) in any randomly initialized, dense neural network \cite{frankle2018lottery}.
Locating these excess parameters via specialisation in different tasks as shown in this work could form the \qquote{natural-selection} counterpart to simple value threshold pruning commonly in use (e.g., see Zhu and Gupta~\cite{zhu2017prune}).


Our approach of utilizing small, randomly initialized networks in a larger meta-structure to predict connectivity results also shares similarity with the network kernel recently introduced by Amid et al.~\cite{amid2022learning}, which returns the expected inner-product of the network logits given some input samples. 
Lin et al.~\cite{lin2013network} also employ a secondary \qquote{micro} network as a function approximator in-between layers, albeit focused on a convolutional network architecture. 
We use a simple architecture, as Tolstikhin et al.~\cite{tolstikhin2021mlp} have shown the sufficient potency of multi-layer perceptron (MLP) architectures for vision tasks.


Finally, with each of our \acl{on} also training themselves independently, we point to the Particle Swarm Optimization (PSO) algorithm (see Carvalho and Ludermir~\cite{carvalho2007particle}) as conceptually related work, which changes and improves its network parameters by iterating through fitness-based points in solution space (particles).

\section{Conclusion and Future Work}
\label{sec:conclusion_and_future_work}

In this work, we built a cooperating \acl{on} made from cells of \qquote{self-replicators} to introduce the field of \ac{srnn}-research to real-world tasks and explore the findings along the way. 
Our approach employs \ac{srnn} as weight approximators for a greater computational graph (\ac{nn}). 
We achieved our goal in training the ON (which is made up of smaller groups of particles) in an end-to-end fashion. 
We show a simple proof-of-concept goal (float-number addition) as well as a non-trivial task (MNIST image classification) to work respectably well.

On our way, we have found that the particles can become robust goal-networks themselves, i.e., \acp{srnn} performing both self-replication and \ac{mnist} classification. 
This dual status becomes harder to reach as \acp{pn} are utilised in more complex group actions.
For the image classification task, we see the emergence of this dual-ability occurring primarily in particles responsible for areas of low importance in the input, signaling the \qquote{leisure} to specialise into both tasks. 

As \acp{pn} represent an approximation of the weight scalar, we can deduct the weight value activating each particle, allowing us to populate conventional feed-forward NN layers.
When pruning parameters on corresponding positions (to \pnsr) of a comparable \ac{nn} we observe minimal test-accuracy loss, which seems to outperform simple `l1-norm' pruning.
For the future, it remains interesting to work on exploring further aspects of learned self-regulation within the organism network, e.g., the usage of particles as learned cell aggregators or layer activation function approximators.
Furthermore, operator-based interaction between particles has to be further explored. 

The option to scale up training and input dimensions for the organism network is also a priority.
As our networks stayed relative simple \wrt{} to the architecture, function, and training regime, there is a lot of methodology from the well-established field of \qquote{deep learning} that we did not touch yet.
Other directions for further research are a more in-depth examination of the viability of different data-sets on the training ability of the organisms.
Although we have shown the importance-selection to work also with a noisy rendition of MNIST, it will be interesting to see the impact of RGB data without any dead-spots (i.e., CIFAR10). 
We leave those considerations as well as further tests and comparisons of mentioned pruning strategies discussed in related work (repeated training, self-pruning, fine-tuning) for the future.


\footnotesize
\bibliographystyle{ieeetr}
\bibliography{bib}

\end{document}